\documentclass[twoside,leqno,twocolumn]{article}

\usepackage[letterpaper]{geometry}

\usepackage{ltexpprt}
\usepackage{hyperref}
\usepackage{graphicx}
\usepackage{caption}
\usepackage{subcaption}
\usepackage{float}
\usepackage{amsmath}
\usepackage{amssymb}
\usepackage{comment}
\usepackage{color}

\usepackage{ulem}
\usepackage{xspace}
\usepackage[usenames,dvipsnames]{xcolor} 
\usepackage[affil-it]{authblk}

\begin{document}

\title{\Large Nearly Optimal Steiner Trees using Graph Neural Network Assisted Monte Carlo Tree Search}
\author{Reyan Ahmed\thanks{abureyanahmed@arizona.edu}, Mithun Ghosh, Kwang-Sung Jun, Stephen Kobourov}
\affil{University of Arizona, Tucson, AZ, USA}

\date{}

\maketitle







\begin{abstract} \small\baselineskip=9pt 
Graph neural networks are useful for learning problems, as well as for combinatorial and graph problems such as the Subgraph Isomorphism Problem and the Traveling Salesman Problem. 
We describe an approach for computing Steiner Trees by combining a graph neural network and Monte Carlo Tree Search. We first train a graph neural network that takes as input a partial solution and  proposes a new node to be added as output. 
This neural network is then used in a Monte Carlo 
search to compute a Steiner tree. The proposed method consistently outperforms the standard 2-approximation algorithm on many different types of graphs and often finds the optimal solution.

\end{abstract}

\section{Introduction}
Graphs arise in many real-world applications that deal with relational information.
Classical machine learning models, such as neural networks and recurrent neural networks, do not naturally handle graphs.
Graph neural networks (GNN) were introduced by  Gori et al.~\cite{gori2005new} in order to better capture graph structures. A GNN is a recursive neural network where nodes are treated as state vectors and the relationships between the nodes are quantified by the edges. Scarselli et al.~\cite{scarselli2008graph} extended the notion of unfolding equivalence that leads to the transformation of the approximation property of feed-forward networks (Scarselli and Tsoi~\cite{scarselli1998universal}) to GNNs.

Many real-world problems are modeled by combinatorial and graph problems that are known to be NP-complete.
GNNs offer an alternative to traditional heuristics and approximation algorithms; indeed the initial GNN model~\cite{scarselli2008graph} was used to approximate solutions to two classical graph problems: subgraph isomorphism and clique detection.


Recent GNN work~\cite{li2018combinatorial,xing2020graph} suggests that combining neural networks and tree search leads to better results than just the neural network alone. Li et al.~\cite{li2018combinatorial} combine a convolutional neural network with tree search to  compute independent sets and other NP-hard problems that are efficiently reducible to the independent set problem. AlphaGo, by Silver et al.~\cite{silver2016mastering}  combines deep convolutional neural networks and Monte Carlo Tree Search (MCTS)~\cite{coulom2006efficient,kocsis2006bandit} to assess Go board positions and reduce the search space. 
Xing et al.~\cite{xing2020graph} build on this combination to tackle the traveling salesman problem (TSP).

Since Xing et al.~\cite{xing2020graph} showed that the AlphaGo framework is effective for TSP, a natural question is whether this framework can be applied to other combinatorial problems such as the Steiner tree problem. Although both TSP and the Steiner tree problem are NP-complete, they are different. First, in the Steiner tree problem we are given a subset of the nodes called {\it terminals} that must be spanned, whereas in TSP all nodes are equivalent. Second, the output of the Steiner tree problem is a tree, whereas the output of TSP is a path (or a cycle). When iteratively computing a TSP solution, the next node to be added can only be connected to the previous one, rather than having to choose from a set of nodes when growing a Steiner tree.
Third, TSP and Go are similar in terms of the length of the instance: both the length of the game and the number of nodes in the TSP solution are fixed and taking an action in Go is equivalent to adding a node to the tour, while the number of nodes in the Steiner tree problem varies depending on the graph instance. Finally, Xing et al.~\cite{xing2020graph} only considered geometric graphs, which is a restricted class of graphs.

\subsection{Background:}
The Steiner tree problem is one of Karp's 21 NP-complete problems~\cite{karp1972reducibility}:
given an edge-weighted graph $G=(V,E)$, a set of terminals $T\subseteq V$ and cost $k$, determine whether there exists a tree of cost at most $k$ that spans all terminals. For $|T|=2$ this is equivalent to the shortest path problem, for $|T|=|V|$ this is equivalent to the  minimum spanning tree problem, while for $2<|T|<|V|$ the problem is NP-complete~\cite{cormen2009introduction}. 
Due to applications in many domains, there is a long history of heuristics,  approximation algorithms and exact algorithms for the problem.
The classical 2-approximation algorithm for the Steiner tree problem~\cite{Gilbert1968}
uses the {\it metric closure} of~$G$, i.e., the complete edge-weighted graph~$G^*$ with terminal node set~$T$ in which, for every edge $uv$, the cost of~$uv$ equals the length of a shortest $u$--$v$ path in~$G$.
A minimum spanning tree of~$G^*$ corresponds to a 2-approximation Steiner tree in~$G$.
This algorithm is easy to implement and performs well in practice~\cite{ahmed2019multi}.
The last in a long list of improvements is the LP-based  algorithm of Byrka et al.~\cite{Byrka2013}, with approximation ratio of $\ln(4)+\varepsilon < 1.39$.
The Steiner tree problem is APX-hard \cite{Bern1989} and NP-hard to approximate  within a factor of $96/95$~\cite{Chlebnik2008}.
Geometric variants of the problem, where terminals correspond to points in the Euclidean or rectilinear plane, admit polynomial-time approximation schemes \cite{Arora1998,Mitchell1999}. 

\begin{figure*}[ht]
    \centering
    \includegraphics[width=1.9\columnwidth]{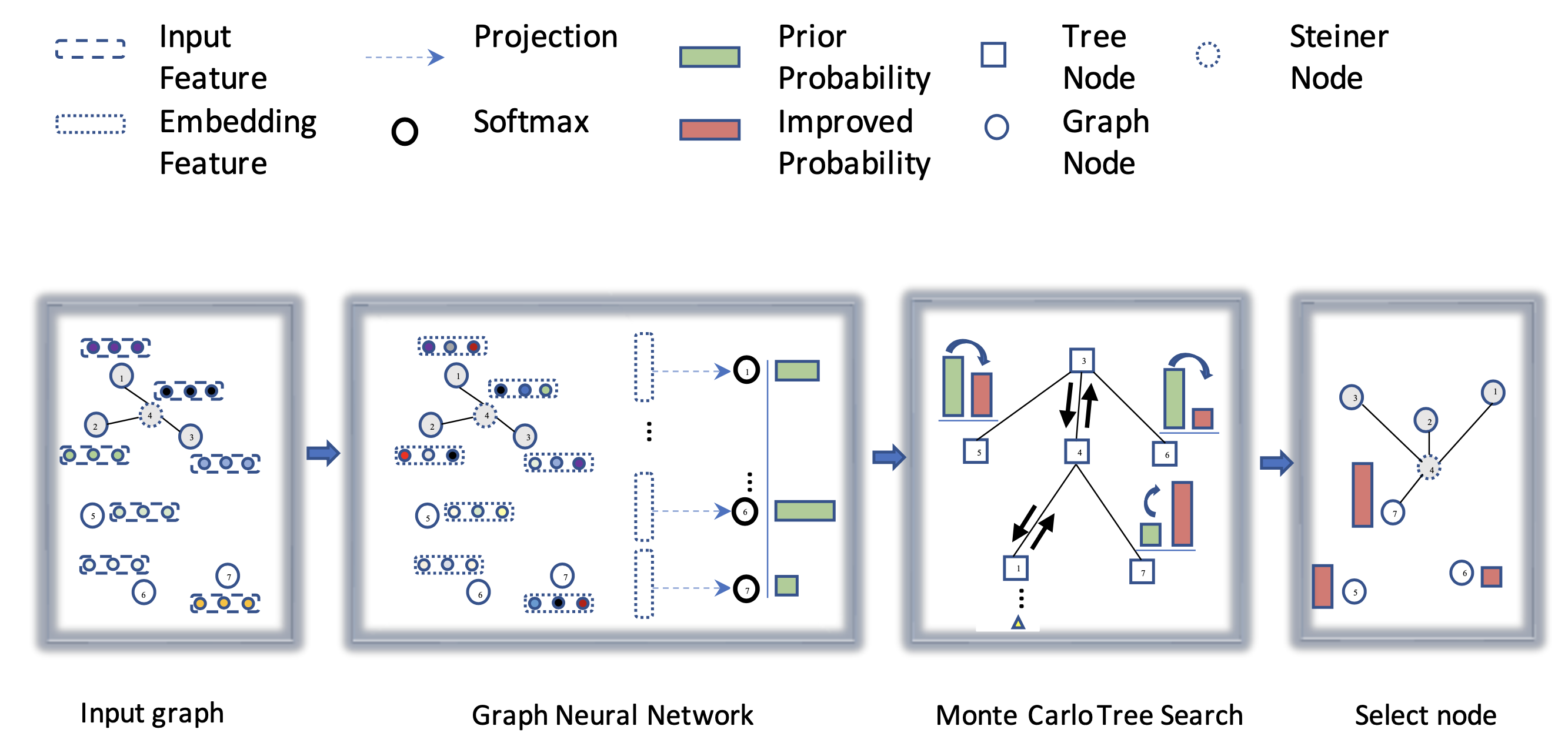}
    \caption{GNN assisted MCTS: first, train a GNN to evaluate non-terminal nodes, then use the network and heuristics to compute a Steiner tree with MCTS.}
    \label{fig:model}
\end{figure*}

\subsection{Related Work:}

Despite its practical and theoretical importance, the Steiner tree problem is not as well explored 
with machine learning approaches as other combinatorial and graph problems. 
In 1985, Hopfield et al.~\cite{hopfield1985neural} proposed a neural network to compute feasible solutions for different combinatorial problems such as TSP. Bout et al.~\cite{den1988traveling} developed a TSP objective function that works well in practice and Brandt et al.~\cite{yao1988alternative} provided different networks for solving TSP. 
Kohonen's 1982 self-organizing maps~\cite{kohonen1982self}, an architecture for artificial neural networks, can also be used for such problems as shown by Fort~\cite{fort1988solving,angeniol1988self} and Favata et al.~\cite{favata1991study}.


Recently, graph neural networks have been an active area of research.  Lei et al.~\cite{lei2017deriving} introduced recurrent neural operations for graphs with associated kernel spaces. Gilmer et al.~\cite{gilmer2017neural} study graph neural models as Message Passing Neural Networks. Garg et al.~\cite{garg2020generalization} generalized message-passing GNNs that rely on the local graph structure, proposing GNN frameworks that rely on graph-theoretic formalisms.  
GNNs have been widely used in many areas including physical systems~\cite{battaglia2016interaction,sanchez2018graph}, protein-protein interaction networks~\cite{fout2017protein}, social science~\cite{hamilton2017inductive,kipf2016semi}, and knowledge graphs~\cite{hamaguchi2017knowledge}; The survey of Zhou et al.~\cite{zhou2018graph} covers  GNN methods and applications in general, and the survey of Vesselinova et al.~\cite{vesselinova2020learning} provides more details on attempts to solve combinatorial and graph problems with neural networks.

\subsection{Problem Statement:}
In the standard optimization version of the Steiner Tree Problem 
we are given a weighted graph $G = (V, E)$ and a set of terminals $T \subseteq V$, and the objective is to compute a minimum cost tree that spans $T$. A Steiner tree $H$ must contain all the terminals and non-terminal nodes in $H$ are the Steiner nodes. Several approximation algorithms have been proposed for this problem including a classical 2-approximation algorithm that first computes the metric closure of $G$ on $T$ and then returns the minimum spanning tree~\cite{agrawal1995trees}. In this paper we 
consider whether graph neural networks can be used to compute spanning trees with close-to-optimal costs using a variety of different graph classes.

\subsection{Summary of Contributions:}

We describe an approach for computing Steiner
Trees by combining a graph neural network and Monte Carlo
Tree Search (MCTS). 
We first train a graph neural network that
takes as input a partial solution and proposes a new node
to be added as output. This neural network is then used
in a MCTS to compute a Steiner tree. The
proposed method consistently outperforms the standard 2-approximation algorithm on many different types of graphs
and often finds the optimal solution.
We illustrate our approach in Figure~\ref{fig:model}.
Our approach builds on the work of Xing et al.~\cite{xing2020graph} for TSP. Since TSP is non-trivially different from the Steiner tree problem, we needed to address challenges in both training the graph neural network and testing the MCTS. We summarize our contribution below:
\begin{itemize}
    \item To train the neural network we generate exact solutions of Steiner tree instances. From each instance, we generate several data points. The purpose of the neural network is to predict the next Steiner node, given a partial solution. Any permutation of the set of Steiner nodes can lead to a valid sequence of predictions. Hence, we use random permutations to generate data points for the network.
    
    \item After we determine the Steiner nodes for a given instance, it is not straightforward to compute the Steiner tree. For TSP, any permutation of all nodes is a feasible tour. For the Steiner tree problem, an arbitrary permutation can have many unnecessary nodes and thus a larger weight compared to the optimal solution. Selecting a subset of nodes is not enough either, since the output needs to be connected and span the terminals. We propose heuristics to compute the tree from the nodes that provide valid result with good quality.
    
    \item We evaluate our results on many different classes of graphs, including geometric graphs, Erd\H{o}s--R\'{e}nyi graphs, Barab\'{a}si--Albert graphs, Watts-Strogatz graphs, and known hard instances from the SteinLib database~\cite{KMV00}. Our method is fully functional and available on Github.
\end{itemize}


\section{Our approach}

Let $G (V, E)$ be a graph, where $V$ is the set of nodes and $E$ is the set of edges. Let $w(u, v)$ be the weight of edge $(u, v) \in E$ and for unweighted graphs $w(u, v) = 1$ for any edge $(u, v) \in E$. Let $T \subseteq V$ be the set of terminals. We use $S = \{v_1, v_2, \cdots, v_i\}$ to represent the set of nodes that are already added in a partially computed Steiner tree. Then, $\overline{S} = V - S$ is the set of candidate nodes to be added to $S$.

Given the graph $G$ our goal is to derive a Steiner tree by adding node $v \in \overline{S}$ to $S$ in turn. A natural approach is to train a neural network to predict which node to add to the partial Steiner tree at a particular step. That is, neural network $f(G|S)$ takes graph $G$ and partial solution $S$ as input, and return probabilities for the remaining nodes, indicating the likelihood they belong to the Steiner tree. 
We use the GNN of~\cite{hamilton2017inductive} to represent $f(G|S)$.

Intuitively, we can directly use the probability values, selecting all nodes with probability higher than a given threshold. We can then  construct a tree from the selected nodes in different ways. For example, we can compute the induced graph of the selected nodes (add an edge if it connects to selected nodes) and extract a minimum spanning tree~\cite{cormen2009introduction}.
Note that the induced graph may be disconnected and therefore the spanning tree will be also disconnected. Even if the spanning tree is connected, it may not span all the terminals, hence it may not provide a valid solution. These issues can be addressed by reducing the given threshold until we obtain a valid solution.

However, deriving trees in this fashion might not be reliable, as a learning-based algorithm has only one chance to compute the optimal solution, and it never goes back to reverse the decision. To overcome this drawback, we leverage the MCTS. 
We use a variant of PUCT~\cite{rosin2011multi} to balance exploration (i.e., visiting a state as suggested by the prior policy) and exploitation (i.e., visiting a state that has the best value). Using the concept of prior probability, the search space of the tree could be reduced substantially, enabling the search to allocate more computing resources to the states having higher values. 
We could get a more reliable policy after a large number of simulations as the output of the MCTS acts as the feedback information by fusing the prior probability with the scouting exploration. The overall approach is illustrated in Figure~\ref{fig:model}.

\subsection{Graph neural network architecture:}

To get a useful neural network, information about the structures of the concerned graph, terminal nodes, and contextual information, i.e., the set of added nodes $S = \{v_1, \hdots , v_i\}$ in the partial solution, is required. We tag node $u$ with $x^t_u=1$ if it is a terminal, otherwise $x^t_u=0$. We also tag node $v$ with $x^a_v=1$ if it is already added, otherwise $x^a_v=0$. Intuitively, $f(G|S)$ should summarize the state of such a ``tagged'' graph and generate the prior probability for each node to get included in $S$.

Some combinatorial problems like the independent set problem and minimum vertex cover problem do not consider edge weights. However, edge weight is an important feature of the Steiner tree problem as the objective is computed based on the weights. Hence, we use the static edge graph neural network (SE-GNN)~\cite{xing2020graph}, to efficiently extract node and edge features of the Steiner tree problem.

A GNN model consists of a stack of $L$ neural network layers, where each layer aggregates local neighborhood information, i.e., features of neighbors of each node, and then passes this aggregated information to the next layer. We use $H_u^l \in \mathbb{R}^d$ to
denote the real-valued feature vector associated with node $u$ at layer $l$. Specifically, the basic GNN model~\cite{hamilton2017inductive} can be implemented as follows. In layer $l = 1, 2, \cdots, L$, a new feature is computed as given by~\ref{eqn:gnn}.

\begin{equation}\label{eqn:gnn}
H_u^{l+1} = \sigma \Big( \theta_1^l H_u^l + \sum_{v \in N(u)} \theta_2^l H_v^l \Big)
\end{equation}

In~\ref{eqn:gnn}, $N(u)$ is the set of neighbors of node $u$, $\theta_1^l$ and $\theta_2^l$ are the parameter matrices for the layer $l$, and $\sigma(\cdot)$ denotes a component-wise non-linear function such as a sigmoid or a ReLU function. For $l=0$, $H_u^0$ denotes the feature initialization at the input layer.

The edge information is not taken into account in~\ref{eqn:gnn}. 
To incorporate edge features, we adapt the approach in~\cite{khalil2017learning,xie2018crystal} to the Steiner tree problem. We integrate the edge features with node features using~\ref{eqn:se-gnn}.

\begin{equation}\label{eqn:se-gnn}
\mu_u^{l+1} = \sigma \Big( \theta_1 x_u + \theta_2 \sum_{v \in N(u)} \mu_v^l + \theta_3 \sum_{v \in N(u)} \sigma(\theta_4 w(u, v)) \Big)
\end{equation}

In~\ref{eqn:se-gnn}, $\theta_1 \in \mathbb{R}^l$, $\theta_2, \theta_3 \in \mathbb{R}^{l\times l}$ and $\theta_4 \in \mathbb{R}^l$ are all model parameters. We can see in~\ref{eqn:gnn} and~\ref{eqn:se-gnn} that the nonlinear mapping of the
aggregated information is a single-layer perceptron, which is not enough to map distinct multisets into unique embeddings. Hence, as suggested in~\cite{xing2020graph,xu2018how}, we replace the single perceptron with a multi-layer perceptron. Finally, we compute a new node feature $H$ using~\ref{eqn:se-gnn2}.

\begin{equation}\label{eqn:se-gnn2}
H_u^{l+1} = \text{MLP}^l \Big( \theta_1^l H_u^l + \sum_{v \in N(u)} \theta_2^l H_v^l + \sum_{v \in N(u)} \theta_3^l e_{u, v} \Big)
\end{equation}

In~\ref{eqn:se-gnn2}, $e_{u, v}$ is the edge feature, $\theta_1^l$, $\theta_2^l$, and $\theta_3^l$ are parameter matrices, and MLP$^l$ is the multi-layer perceptron for  layer $l$. Note that SE-GNN differs from GEN~\cite{dai2017learning} in the following aspects: (1) SE-GNN replaces $x_u$ in~\ref{eqn:se-gnn} with $H_u$ so that the SE-GNN can integrate the latest feature of the node itself directly. (2) Each update process in the GEN can be treated as one update layer of the SE-GNN, i.e., each calculation is equivalent to going one layer forward, thus calculating $L$ times for $L$ layers. Parameters of each layer of SE-GNN are independent, while parameters in GEN are shared between different update processes which limits the neural network. (3) We replace $\sigma$ in~\ref{eqn:se-gnn} with MLP as suggested by~\cite{xing2020graph,xu2018how} to map distinct multisets to unique embeddings.

We initialize the node feature $H^0$ as follows. Each node has a feature tag which is a $4$-dimensional vector. The first element of the vector is binary and it is equal to 1 if the partial solution $S$ contains the node. The second element of the vector is also binary and it is equal to 1 if the node is a terminal. The third and fourth elements of the feature tag are the $x$ and $y$ coordinates of the node. The last two are used only for geometric graphs.

\subsection{Parameterizing $f(G|S; \theta)$:}

Once the feature for every node is computed after updating $L$ layers, we use the new feature for the nodes to define the $f(G|S; \theta)$ function, which returns the prior probability for each node indicating how likely the node will belong to partial solution $S$. Specifically, we fuse all node feature $H_u^L$ as the current state representation of the graph and parameterize $f(G|S; \theta)$ as expressed by~\ref{eqn:se-gnn-out}.

\begin{equation}\label{eqn:se-gnn-out}
f(G|S; \theta) = \text{softmax}( sum(H_1^L) , \cdots , sum(H_n^L) ) 
\end{equation}

During training, we minimize the cross-entropy loss for each training sample $(G_i , S_i )$ in a supervised manner as given by~\ref{eqn:se-gnn-loss}.

\begin{equation}\label{eqn:se-gnn-loss}
\ell(S_i, f(G_i|S_i; \theta)) = - \sum_{j=1}^N y_j \log f(G_i|S_i(1:j-1); \theta)  
\end{equation}

In~\ref{eqn:se-gnn-loss}, $S_i$ is an ordered set of nodes of a partial solution which is a permutation of the nodes of graph $G_i$, with $S_i(1:j-1)$ the ordered subset containing the first $j-1$ elements of $S_k$, and $y_j$ a 
vector of length $N$ with 1 in the $S_i(j)$-th position. We provide more details in Section~\ref{sec:training}.

\subsection{GNN assisted MCTS:}

Similar to the implementation in~\cite{xing2020graph}, the GNN-MCTS uses graph neural networks as a guide of MCTS. We denote the child edges of $u$ in MCTS by $A(u)$. Each node $u$ in the search tree contains edges $(u, a)$ for all legal actions $a \in A(u)$. Each edge of MCTS stores a set of statistics:
$$\{N(u, a), Q(u, a), P(u, a)\},$$
where node $u$ denotes the current state of the graph including the set of nodes $S$ and other graph information, action $a$ denotes the selection of node $v$ from $\overline{S}$ to add in $S$, $N(u, a)$ is the visit count, $Q(u, a)$ is the action value and $P(u, a)$ is the prior probability of selecting edge $(u, a)$.

In the Steiner tree problem, we are interested in finding a tree with minimum cost. Hence, we track the best action value found under the subtree of each node to determine the ``exploitation value" of the tree node, as suggested in~\cite{gao2019tree} in the context of the stock trading problem.

The standard MCTS takes solution values in the range $[0, 1]$~\cite{kocsis2006bandit}. However, the Steiner tree can have an arbitrary solution value that does not fall in a predefined interval. This issue could be addressed by adjusting the parameters of the tree search algorithm in such a way that it is feasible for a specified interval. Adjusting parameters requires substantial trial and error due to the change in the number of nodes. Instead, we address this issue by normalizing the action value of node $n$, whose parent is node $p$, in the range of $[0, 1]$ using~\ref{eqn:MC-normalize}.

\begin{equation}\label{eqn:MC-normalize}
Q_n = \frac{\tilde{Q}_n - w_p}{b_p-w_p}  
\end{equation}

\begin{figure}[ht]
    \centering
    \includegraphics[width=.6\linewidth]{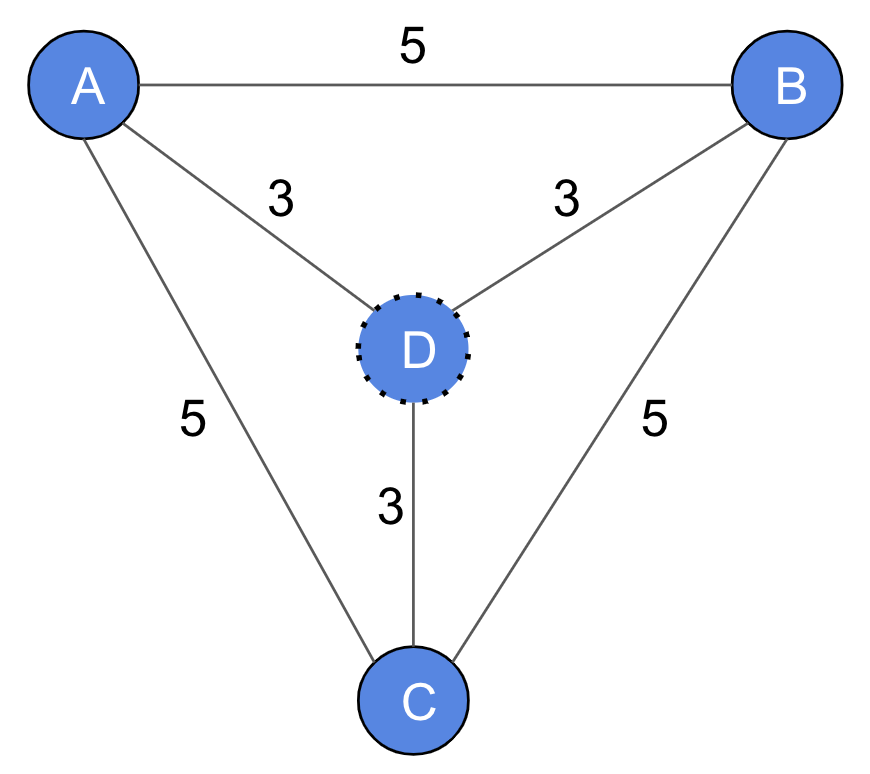}
    \caption{Example graph for the Steiner tree heuristic. Considering $D$ as a terminal node and computing the MST on the metric closure provides a better solution than the 2-approximation.}
    \label{fig:heuristic}
\end{figure}

In~\ref{eqn:MC-normalize}, $b_p$ and $w_p$ are the minimum and maximum action values among the children of $p$, and $Q_n$ is the action value of $n$. The actions under $p$ are normalized in the range of $[0, 1]$ so that the best action is 0 and the worst action is 1.

The GNN-MCTS proceeds by iterating over the four phases below and then selects a move to play.\begin{enumerate}
    \item \textbf{Selection Strategy.} The first in-tree phase of each simulation starts at the root node $v_0$ of the search tree and finishes when the simulation reaches a leaf node $v_l$ at time step $l$. At time step $t<l$, we use a variant of PUCT~\cite{rosin2011multi} to balance exploration (i.e., visiting the states suggested by the prior policy) and exploitation (i.e., visiting states which have best values) according to the statistics in the search tree as given by~\ref{eqn:action} and~\ref{eqn:uncertainity} respectively.
    \begin{equation}\label{eqn:action}
        a_t = \text{arg} \text{max}_a(Q(v_t, a) + U(v_t, a))  
    \end{equation}
    \begin{equation}\label{eqn:uncertainity}
        U(v, a) = c_{puct}P(v, a)\frac{\sqrt{\sum_b N(v, b)}}{1+N(v, a)}
    \end{equation} where $c_{puct}$ is a constant for trading off between exploration and exploitation. We set $c_{puct}=1.3$ according to previous experimental results~\cite{xing2020graph}.
    \item \textbf{Expansion Strategy.} When a leaf node $v$ is reached, the corresponding state $s_v$ is evaluated by the GNN to obtain the prior probability $p$ of its children nodes. The leaf node is expanded and the statistic of each edge $(s_v, a)$ is initialized to $\{N(s_v, a) = 0, Q(s_v, a) = -\infty, P(s_v, a) = p_a\}$.
    \item \textbf{Back-Propagation Strategy.} For each step $t < l$, the edge statistics are updated in a backward process. The visit counts are increased as $N(v_t, a_t) = N(v_t, a_t) + 1$, and the action value is updated to the best value.
    \item \textbf{Play.} After repeating steps 1-3 several times (800 times for smaller datasets and 1200 times for larger datasets according to the previous experimental results~\cite{xing2020graph}), we select the node with the biggest $\hat{P}(a|u_0) = \frac{\tilde{Q}(u_0, a)}{\sum_b \tilde{Q}(u_0, b)}$ as the next move $a$ in the root position $u_0$. The selected child becomes the new root node and the statistics stored in the subtree are preserved.
\end{enumerate}

\subsection{Computing Steiner tree from $S$:\label{sec:heuristics}}
There are several ways to compute a Steiner tree from the set of nodes $S$. We provide two effective heuristics that we use in our experiments.\begin{enumerate}
    \item \textbf{MST-based heuristic.} In this heuristic, we first add the terminal nodes to the solution if they are not already present, and then compute the induced graph. 
    We iteratively add nodes from $\overline{S}$ in order computed by the MCTS until the induced graph is connected. 
    In the last step, we compute a minimum spanning tree (MST) of the induced graph and prune degree-1 non-terminal nodes. 
    This heuristic is effective for geometric graphs and unweighted graphs.
    \item \textbf{Metric closure-based heuristic.} In this heuristic, given an input graph $G = (V, E)$ and a set of terminals $T$, we first compute a metric closure graph $G' = (T, E')$. Every pair of nodes in $G'$ is connected by an edge with weight equal to the shortest path distance between them.   
    The minimum spanning tree of the metric closure provides a 2-approximation to the optimal Steiner tree. 
    For example, in Figure~\ref{fig:heuristic}, $A$, $B$ and $C$ are terminal nodes and $D$ is not. 
    Note that $D$ does not appear in any shortest path as every shortest path between pairs of terminals is 5 and none of them goes through $D$. 
    Without loss of generality, the $2$-approximation algorithm chooses the $A-C-B$ path with  total cost of 10, while the optimal solution that uses $D$ has cost 9. 
    
    While the $2$-approximation algorithm does not consider any node that does not belong to a shortest path between two terminal nodes, here we consider such nodes. Specifically,
    we iteratively add nodes from $\overline{S}$ in order computed by the MCTS, even if they don't belong to any shortest path.
    Note that, unlike the MST-based heuristic, the metric closure-based heuristic computes the MST on the metric closure (not on the input graph).
\end{enumerate}
Both of the heuristics start by selecting all the terminals as the partial solution. In the MCTS, we gradually add nodes that are not in the set of already selected nodes. For the MST-based heuristic, we stop selecting nodes when the induced graph becomes connected. For the metric closure-based heuristic we stop selecting nodes when $10\%$ non-terminal nodes have been selected.

\begin{figure*}[ht]
\minipage{0.46\textwidth}
  \includegraphics[width=\linewidth]{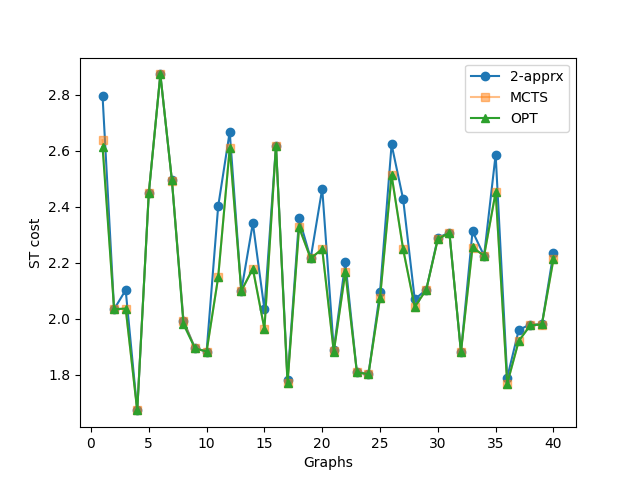}
  \subcaption{Random geometric graphs with 20 nodes}
  \label{fig:GE}
\endminipage\hfill
\minipage{0.46\textwidth}
  \includegraphics[width=\linewidth]{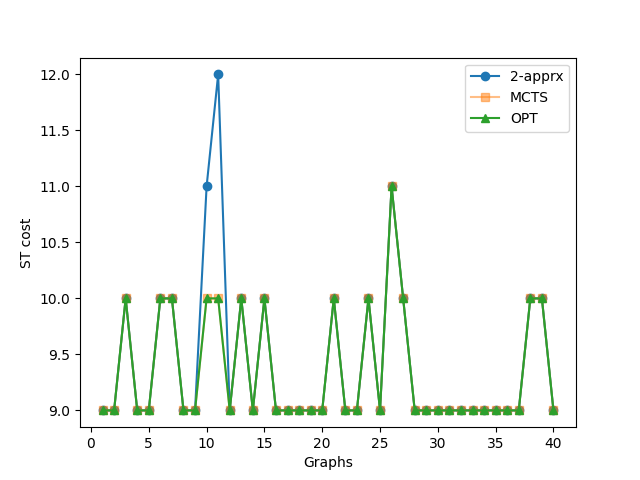}
  \subcaption{Unweighted Erd\H{o}s--R\'{e}nyi graphs with 20 nodes}
  \label{fig:ER}
\endminipage\hfill
\minipage{0.46\textwidth}%
  \includegraphics[width=\linewidth]{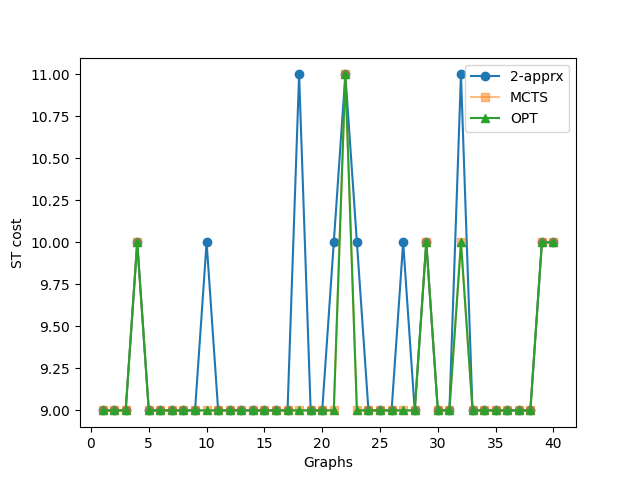}
  \subcaption{Unweighted Barab\'{a}si--Albert graphs with 20 nodes}
  \label{fig:BA}
\endminipage\hfill
\minipage{0.46\textwidth}%
  \includegraphics[width=\linewidth]{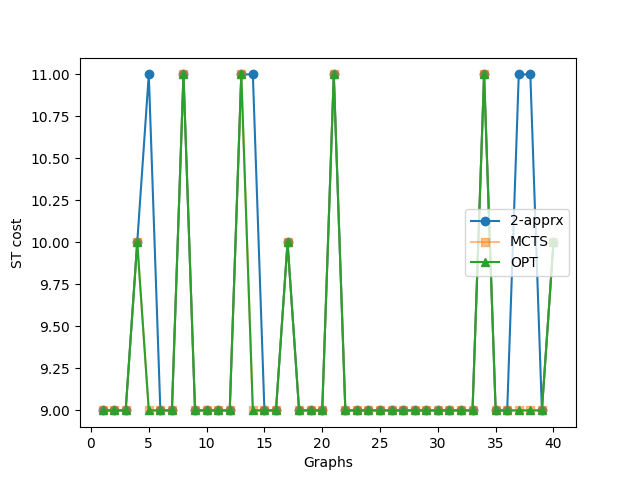}
  \subcaption{Unweighted Watts--Strogatz graphs with 20 nodes}
  \label{fig:WS}
\endminipage
\caption{Performance on simple graphs. Each data point represents one graph. The lower the cost the better the algorithm is. Our algorithm (MCTS) is nearly optimal and performs better than 2-approximation.}
\label{fig:performance_simple}
\end{figure*}

\section{Model setup and training}\label{sec:training}

In order to train the models, one has to provide training data consisting of input graphs $G = (V, E)$, edge weights $W:E \rightarrow \mathbb{R}^+$, and terminals $T \subseteq V$. Given $G, W, T$, and partial solution $S$, our goal is to give label 1 to the next node to be added and 0 to all others. 
Initially, we set $S=T$ as all terminals must be in the Steiner tree. Consider a graph with 6 nodes $u_1, u_2, \cdots, u_6$, $T=\{u_1, u_2, u_3\}$, and an optimal Steiner tree contains the first five nodes $u_1, u_2, \cdots, u_5$. For this example, initially we set $S = T = \{u_1, u_2, u_3\}$. Since we have two Steiner nodes $u_4$ and $u_5$, both permutations $u_4, u_5$ and $u_5, u_4$ are valid. For the first permutation, after setting $S=\{u_1, u_2, u_3\}$, the next node to be added in the solution is $u_4$. Hence for this data point, only the label for $u_4$ is 1. This permutation provides another data point where $S = \{u_1, u_2, u_3, u_4\}$ and only the label for $u_5$ is equal to 1. Similarly, we can generate two more data points from the other permutation. This exhaustive consideration of all possible permutations does not scale to larger graphs, so we randomly select 100 permutations from each optimal solution. The model is trained with Stochastic Gradient Descent, using the ADAM optimizer~\cite{kingma2014adam} to minimize the cross-entropy loss between the models' prediction and the ground-truth (a 
vector in $\{0, 1\}^{|V|}$ indicating whether a node is the next solution node or not) for each training sample.

\subsection{Data generation:}
We produce training instances using several different random graph generation models:  Erd\H{o}s--R\'{e}nyi~\cite{erdos1959random}, Watts--Strogatz~\cite{watts1998collective}, Barab\'{a}si--Albert~\cite{barabasi1999emergence}, and random geometric~\cite{penrose2003random} graphs. Each of these generators needs some parameters; below we describe the values we used, aiming to have graphs of comparable density across the different generators. For Erd\H{o}s--R\'{e}nyi model, there is an edge selection probability $p$, which we set to $\frac{2\ln n}{n}$ to ensure that the generated graphs are connected with high probability. In the Watts--Strogatz model, we initially create a ring lattice of constant degree $K$ and rewire each edge with probability $0 \leq p \leq 1$, while avoiding self-loops and duplicate edges. For our experiments we use $K=6$ and $p=0.2$. 
In the Barab\'{a}si--Albert model, the graph begins with an initially connected graph of $m_0$ nodes. New nodes are added to the network one at a time. Each new node is connected to $m \leq m_0$ existing nodes with a probability that is proportional to the number of edges that the existing nodes already have. We set $m_0=5$.
In the random geometric graph model, we uniformly select $n$ points from the Euclidean cube, and connect nodes whose Euclidean distance is not larger than a threshold $r_c$, which we choose to be $\sqrt{\frac{2\ln n}{\pi n}}$ to ensure the graph is connected with high probability.

The Steiner tree problem is NP-complete even if the input graph is unweighted~\cite{garey1979computers}. We generate both unweighted and weighted Steiner tree instances using the random generators described above. The number of nodes in these instances is equal to 20 and the number of terminals is equal to 10.
For each type of instance we generate 200 instances.
For weighted graphs, we assign random integer weights in the range $\{1, 2, \cdots, 10\}$ to each edge. Since the weighted version of the Steiner tree problem is the more general version, and the number of terminals is an important parameter, we create a second dataset of graphs with 50 nodes. For the number of terminals, we use two distributions. In the first distribution, the percentage of the number of terminals with respect to the total number of nodes is in $\{20\%, 40\%, 60\%, 80\%\}$. In the second distribution the percentage is in $\{3\%, 6\%, \cdots, 18\%\}$. These two cases are considered to determine the behavior of the learning models on large and small terminal sets (compared with the overall graph size).
As random graphs instances can be ``easy" to solve, we also evaluate our approach on graphs from the SteinLib library~\cite{KMV00}, which provides hard graph instances.
Specifically, we perform experiments on two SteinLib datasets: \href{http://steinlib.zib.de/showset.php?I080}{I080} and \href{http://steinlib.zib.de/showset.php?I160}{I160}.
Each instance of the I080 and I160 datasets contains 80 nodes and 160 nodes respectively. Both datasets have 100 instances.

\subsection{Computing optimal solutions:}
In order to evaluate the performance of our approach (and that of the 2-approximation) we need to compute the optimal solutions.
There are different integer linear programs (ILP) for the exact Steiner tree problem. The cut-based approach considers all possible combinations of partitions of terminals and ensures that there is an edge between that partition. This ILP is simple but introduces an exponential number of constraints. A better ILP approach in practice considers an arbitrary terminal as a root and sends flow to the rest of the terminals; see~\cite{ahmed2019multi,jabrayilov2019new} for details about these and other ILP methods for the exact Steiner tree problem.

We  generate 2,000 Steiner tree instances and compute the exact solution with the flow-based ILP. We use CPLEX 12.6.2 as the ILP solver on a high-performance computer (Lenovo NeXtScale nx360 M5 system with 400 nodes with 192 GB of memory each). We  use Python to implementing the algorithms described above.

\begin{figure*}[ht]
\minipage{0.46\textwidth}
  \includegraphics[width=\linewidth]{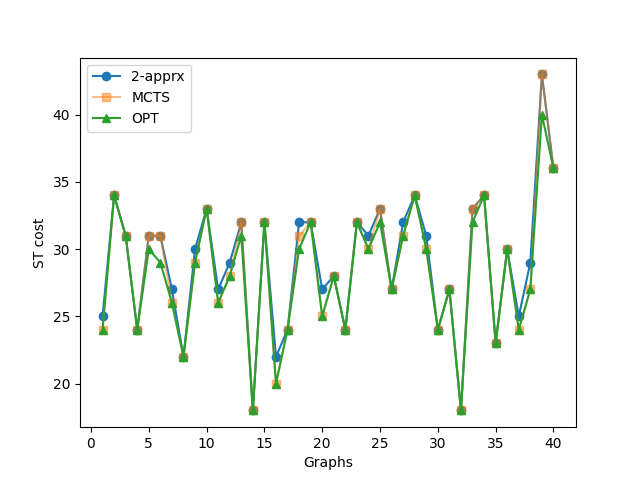}
  \subcaption{Weighted Erd\H{o}s--R\'{e}nyi graphs with 20 nodes}
  \label{fig:ER20}
\endminipage\hfill
\minipage{0.46\textwidth}
  \includegraphics[width=\linewidth]{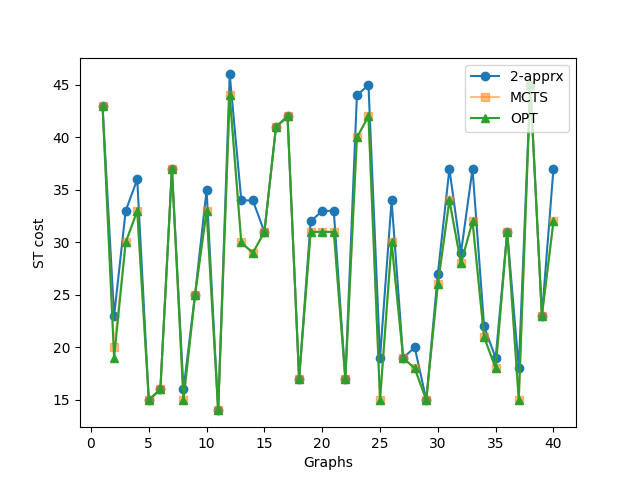}
  \subcaption{Weighted Barab\'{a}si--Albert graphs with 20 nodes}
  \label{fig:BA20}
\endminipage\hfill
\minipage{0.46\textwidth}%
  \includegraphics[width=\linewidth]{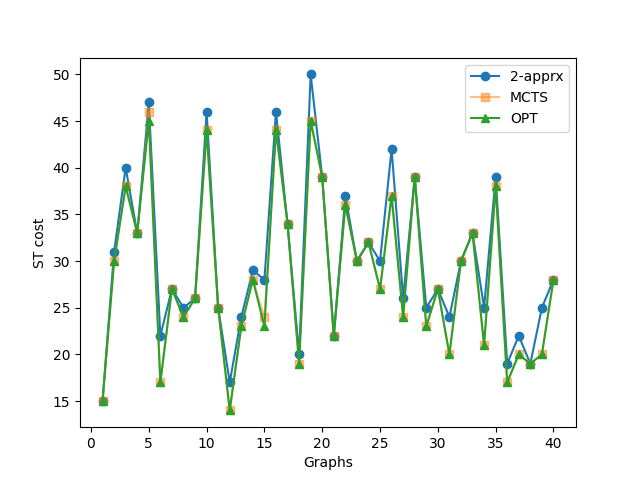}
  \subcaption{Weighted Watts-Strogatz graphs with 20 nodes}
  \label{fig:WS20}
\endminipage\hfill
\minipage{0.46\textwidth}%
  \includegraphics[width=\linewidth]{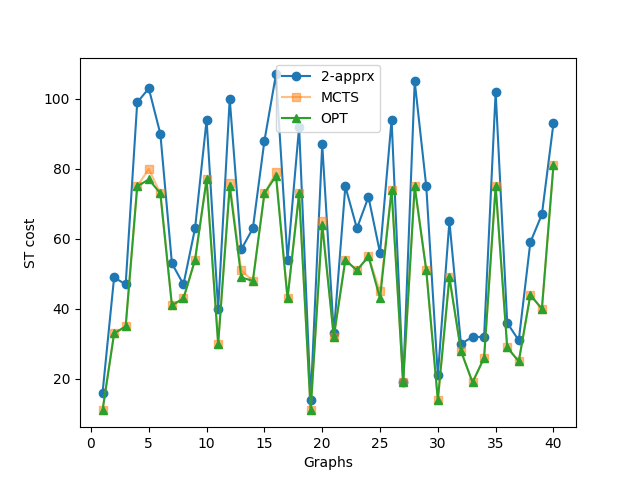}
  \subcaption{Weighted Erd\H{o}s--R\'{e}nyi graphs with 50 nodes}
  \label{fig:ER50}
\endminipage\hfill
\caption{Performance on weighted graphs. Each data point represents one graph. The lower the cost the better the algorithm is. Our algorithm (MCTS) is nearly optimal and performs better than 2-approximation.}
\label{fig:performance_complex}
\end{figure*}

\subsection{Model architectures:}
For the MLP in our GNN model, we have used two hidden layers. The first hidden layer has an embedding dimension equal to 128. The second hidden layer has a convolution dimension equal to 128. We use the ReLU activation function for both layers. We also use batch normalization in both layers to normalize the contributions to a layer for every batch of the datasets. The value of early stopping is equal to 15; hence the model will automatically stop training when the chosen metric does not improve for 15 epochs.
We trained the network and evaluated our algorithm separately for each combination of generator and node size. Recall that the neural network predicts the next Steiner node from a partial solution. Hence, for each Steiner tree instance, we generate a set of data points. {\color{black}Since neural network architecture can not handle different node sizes, we have trained four independent neural networks for node sizes 20, 50, 80, and 160. The same neural network can predict solution nodes for different graph generation models if the node size is the same.} In total, we have trained the networks on around 200,000 data points.

\subsection{Heuristic setup:}
We used the two heuristics described in Section~\ref{sec:heuristics}.  
Recall that the MST-based heuristic just computes the minimum spanning tree on the induced graph of the partial solution. It works well for geometric graphs, unweighted Erd\H{o}s--R\'{e}nyi, unweighted Watts--Strogatz, and unweighted Barab\'{a}si--Albert graphs. 
We use the metric closure-based heuristic for all the other experiments.

\begin{figure*}[ht]
\minipage{0.46\textwidth}%
  \includegraphics[width=\linewidth]{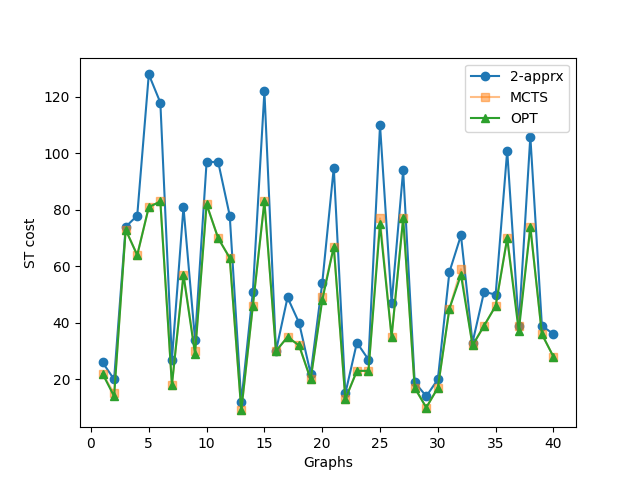}
  \subcaption{Weighted Watts-Strogatz graphs with 50 nodes}
  \label{fig:WS50}
\endminipage\hfill
\minipage{0.46\textwidth}%
  \includegraphics[width=\linewidth]{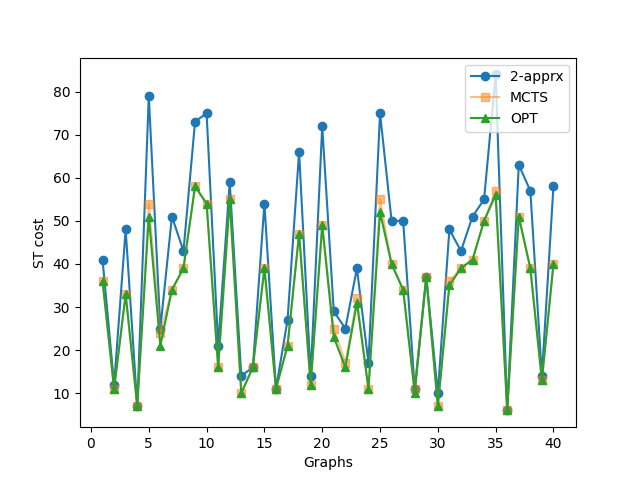}
  \subcaption{Weighted Barab\'{a}si--Albert graphs with 50 nodes}
  \label{fig:BA50}
\endminipage\hfill
\caption{Performance on more weighted graphs. Each data point represents one graph. The lower the cost the better the algorithm is. Our algorithm (MCTS) is nearly optimal and performs better than 2-approximation.}
\label{fig:performance_complex_part2}
\end{figure*}

\section{Experimental results}\label{sec:experiment}

We evaluate the performance of the proposed approach by comparing the computed trees to those computed by the classical 2-approximation algorithm and the optimal solutions. The proposed approach never performs worse than the 2-approximation algorithm. We also report running times. 

The results for geometric graphs and other unweighted graphs are shown in Figure~\ref{fig:performance_simple}.
{\color{black} The X-axis represents the graph or instance number that does not have any significance. Traditionally bar plot is used in such a scenario. However, for each instance, we show three costs for three different algorithms. Hence scatter plot provided a better visualization by saving space horizontally. One can show two costs instead of three costs by showing the difference w.r.t. the optimal algorithm. However, this approach does not provide a better visualization since many differences get closer to zero.}
We illustrate the performance of different algorithms on the Geometric graphs in Figure~\ref{fig:GE}. We represent the optimal solution with green triangles, our algorithm with yellow squares, and the 2-approximation with blue circles. For the geometric graph, we have 40 instances, each of which has 20 nodes and 10 terminals. A majority of the time the 2-approximation has a larger solution value and our algorithm has a solution very close to the optimal value. The 2-approximation performs worse than our algorithm in 36 instances out of 40 instances.

Our algorithm also performs well for unweighted graphs. We illustrate the performance for random graphs generated by Erd\H{o}s--R\'{e}nyi, Barab\'{a}si--Albert, and Watts--Strogatz models in Figure~\ref{fig:ER}, Figure~\ref{fig:BA}, and Figure~\ref{fig:WS} respectively. We have 40 instances for each type of generator. Again, each instance has 20 nodes and 10 terminals. In all of these instances, our algorithm achieves the optimal solution. For Erd\H{o}s--R\'{e}nyi graphs, our algorithm performs better than the 2-approximation in two instances. For Barab\'{a}si--Albert graphs, our algorithm performs better in six instances. For Watts--Strogatz graphs, our algorithm performs better in four instances.

\begin{figure*}[ht]
\minipage{0.46\textwidth}%
  \includegraphics[width=\linewidth]{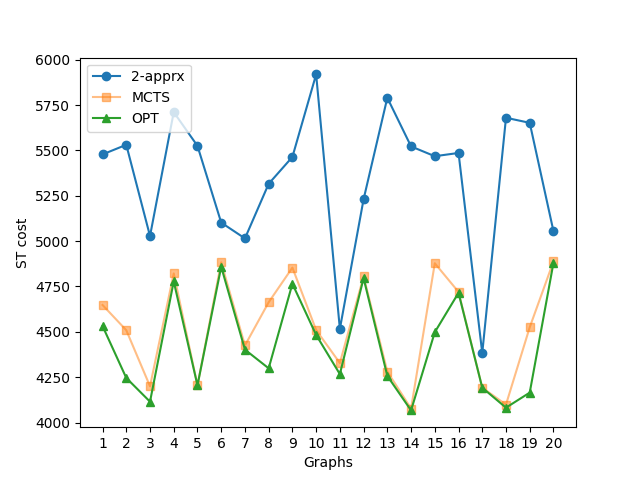}
  \subcaption{Weighted  80 nodes graphs of SteinLib I080 dataset}
  \label{fig:I080}
\endminipage\hfill
\minipage{0.46\textwidth}%
  \includegraphics[width=\linewidth]{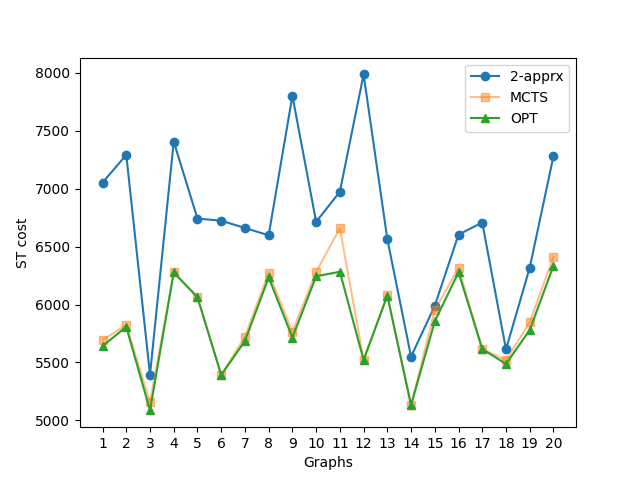}
  \subcaption{Weighted  160 nodes graphs of SteinLib I160 dataset}
  \label{fig:I160}
\endminipage\hfill
\caption{Performance on SteinLib datasets. Each data point represents one graph. The lower the cost the better the algorithm is. Our algorithm (MCTS) is nearly optimal and performs better than 2-approximation.}
\label{fig:performance_SteinLib}
\end{figure*}

\begin{table*}[ht]
    \begin{center}
    \begin{tabular}{|p{1.8cm}|c|c|c|c|c|c|c|c|c|c|c|c|}
    \hline
     Graphs/ Algorithms & GE & ER & WS & BA & ER20 & WS20 & BA20 & ER50 & WS50 & BA50 & I080 & I060 \\ 
     \hline
     2-apprx & 0.16 & 0.09 & 0.10 & 0.11 & 0.16 & 0.40 & 0.14 & 1.14 & 0.79 & 0.47 & 1.29 & 7.88 \\  
     \hline
     MCTS & 0.64 & 0.40 & 0.49 & 0.49 & 0.75 & 1.73 & 0.66 & 5.06 & 3.20 & 2.17 & 5.77 & 34.52 \\  
     \hline
     OPT & 5.92 & 6.33 & 5.00 & 4.68 & 22.99 & 28.61 & 29.90 & 153.71 & 125.41 & 134.46 & 1051.51 & 6188.18 \\  
     \hline
    \end{tabular}
    \end{center}
    \vspace{-.5cm}
    \caption{Average running time of different algorithms in seconds.}
    \label{fig:run_time}
\end{table*}

Results for the weighted graphs are shown in Figure~\ref{fig:performance_complex}. The weighted version of the Steiner tree problem is harder than the unweighted version. Hence, we consider a larger set of instances. For each random graph generation model, we consider one dataset that has 20 nodes for each instance and another dataset that has 50 nodes. We illustrate the performance on 20 nodes Erd\H{o}s--R\'{e}nyi graphs in Figure~\ref{fig:ER20}. For this dataset, both of the algorithms provide solution values similar to the optimal value. We illustrate the performance on 20 nodes Barab\'{a}si--Albert and Watts-Strogatz graphs in Figrue~\ref{fig:BA20} and Figure~\ref{fig:WS20} respectively. For Barab\'{a}si--Albert graphs, the 2-approximation performs worse than our algorithm in 24 instances out of 40 instances. Our algorithm provides an optimal solution in 39 instances. For Watts-Strogatz graphs, the 2-approximation performs worse than our algorithm in 24 instances out of 40 instances. Our algorithm provides an optimal solution in 38 instances.

We illustrate the performance of the algorithms on 50 nodes Erd\H{o}s--R\'{e}nyi graphs in Figure~\ref{fig:ER50}. We can see a larger difference between our algorithm and the 2-approximation for this 50-nodes dataset. We can see that our algorithm is providing nearly optimal solutions. On the other hand, the 2-approximation often provides higher solution values. The 2-approximation performs worse than our algorithm in 39 instances out of 40 instances. Our algorithm provides an optimal solution in 34 instances.

We illustrate the performance of the algorithms on 50 nodes Watts-Strogatz graphs in Figure~\ref{fig:WS50}. Again, our algorithm provides nearly optimal solutions and the 2-approximation has a noticeable difference. The 2-approximation performs worse than our algorithm in 38 instances out of 40 instances. Our algorithm provides an optimal solution in 34 instances. We illustrate the performance of the algorithms on 50 nodes Barab\'{a}si--Albert graphs in Figure~\ref{fig:BA50}. The 2-approximation performs worse than our algorithm in 34 instances out of 40 instances.
Our algorithm provides an optimal solution in 31 instances. Our algorithm provides nearly optimal solutions for the remaining instances.

The SteinLib library~\cite{KMV00} provides hard graph instances for solving the Steiner tree problem. Results for a SteinLib dataset is shown in Figure~\ref{fig:performance_SteinLib}. We can see that there is a relatively large difference between the optimal solution value and the 2-approximation solution value. Despite this larger difference of 2-approximation solution values, our algorithm finds nearly optimal solutions.

\subsection{Running time:}
The training time of the GNN depends on the dataset. The maximum training time is around 20 hours for the I160 SteinLib dataset. The average running times of the optimal algorithm, 2-approximation, and our algorithm for different test datasets are shown in Figure~\ref{fig:run_time}. We denote the geometric, unweighted Erd\H{o}s--R\'{e}nyi, unweighted Watts--Strogatz, and unweighted Barab\'{a}si--Albert graphs by GE, ER, WS, and BA respectively. We denote the weighted 20 nodes Erd\H{o}s--R\'{e}nyi, Watts--Strogatz, and Barab\'{a}si--Albert graphs by ER20, WS20, and BA20 respectively. We denote the weighted 50 nodes Erd\H{o}s--R\'{e}nyi, Watts--Strogatz, and Barab\'{a}si--Albert graphs by ER50, WS50, and BA50 respectively. We denote the 80 nodes and 160 nodes SteinLib datasets by I080 and I160 respectively. We can see the 2-approximation algorithm is the fastest. Our algorithm is a little slower, however, the solution values are closer to the optimal values. 

\section{Conclusion}
We described an approach for the Steiner tree problem based on GNNs and MCTS. 
An experimental evaluation shows that the proposed method computes nearly optimal solutions on a wide variety of datasets in a reasonable time. The proposed method never performs worse than the standard 2-approximation algorithm. The source code and experimental data can be found on github 
{\small \url{https://github.com/abureyanahmed/GNN-MCTS-Steiner}}.

One limitation of our work is we need to retrain for different node sizes. Also, the Steiner tree problem can be seen as a network sparsification technique. In fact, it is one of the simplest sparsification methods since it only considers trees. It would be interesting to see whether our proposed approach can be adapted to  graph spanner problems.
{\color{black}Our model is unable to fit different node sizes. Hence in our experiments, we try a small set of node sizes. It is an interesting future work to generalize the model to handle different node sizes. A general model will provide an opportunity to explore the effectiveness of different parameters of the model.}

\bibliographystyle{plain}
\bibliography{references}

\end{document}